\def\year{2021}\relax
\documentclass[letterpaper]{article} 
\usepackage{aaai21}  
\usepackage{times}  
\usepackage{helvet} 
\usepackage{courier}  
\usepackage[hyphens]{url}  
\usepackage{graphicx} 
\usepackage{natbib}  
\usepackage{caption} 
\usepackage{amsmath}
\usepackage{amsthm}
\usepackage{proof}
\newtheorem{theorem}{Theorem}
\usepackage{comment}

\newcommand{\MG}[1]{{\color{black}{#1}}}

\usepackage[utf8]{inputenc} 
\usepackage[T1]{fontenc}    
\usepackage{hyperref}       
\usepackage{url}            
\usepackage{booktabs}       
\usepackage{amsfonts}       
\usepackage{nicefrac}       
\usepackage{microtype}      
\usepackage{dsfont}
\usepackage{bm}
\usepackage{mathrsfs}
\usepackage{amsmath,mathtools}
\usepackage{amsfonts}
\usepackage{amsthm}
\usepackage{amssymb}
\usepackage{xcolor}
\usepackage{setspace}
\usepackage{float}
\usepackage{url}
\usepackage{multicol}
\usepackage{subfigure}
\usepackage[]{algorithm2e}

\SetCommentSty{mycommfont}

\usepackage{soul}
\mathtoolsset{showonlyrefs}

\title{Partition of unity networks: deep hp-approximation}
\author{
Kookjin Lee\textsuperscript{\rm 1},
Nathaniel A. Trask\textsuperscript{\rm 2},
Ravi G. Patel\textsuperscript{\rm 2},
Mamikon A. Gulian\textsuperscript{\rm 2},
Eric C. Cyr\textsuperscript{\rm 2}
}
\affiliations{

    \textsuperscript{\rm 1} Extreme Scale Data Science \& Analytics Department, Sandia National Laboratories\\
    \textsuperscript{\rm 2}Center for Computing Research, Sandia National Laboratories\\
    Albuquerque, New Mexico, 87123\\
    \{koolee, natrask, rgpatel, mgulian, eccyr\}@sandia.gov

}

\begin{document}

\maketitle

\begin{abstract}
Approximation theorists have established best-in-class optimal approximation rates of deep neural networks by utilizing their
ability to simultaneously emulate partitions of unity and monomials. Motivated by this, we propose partition of unity networks (POUnets) which incorporate these elements directly into the architecture. Classification architectures \MG{of the type used to learn} probability measures are used to build a meshfree partition of space, while polynomial spaces with learnable coefficients are \MG{associated} to each partition. The resulting $hp$-element-like approximation allows use of a fast least-squares optimizer, and the resulting architecture size \MG{need} not scale exponentially with spatial dimension, breaking the curse of dimensionality. An abstract approximation result establishes desirable properties to guide network design. Numerical results for two choices of architecture demonstrate \MG{that POUnets yield} $hp$-convergence for smooth functions and consistently outperform MLPs for piecewise polynomial functions with large numbers of discontinuities.
\end{abstract}

\section{Overview}

We consider regression over the set $\mathcal{D} = \left\{(\mathbf{x}_i,{y}_i)\right\}_{i=1}^{N_\text{data}}$, where $\mathbf{x}_i \in \mathbb{R}^d$ and ${y}_i={y}(\mathbf{x}_i)$ are point samples of a piecewise smooth function. Following success\MG{es} in classification problems \MG{in high-dimensional spaces \cite{francois2017deep}}, deep neural networks (DNNs) have garnered tremendous interest  as tools for regression problems and numerical analysis, partially due to their apparent ability to alleviate the \textit{curse of dimensionality} in the presence of \MG{latent} low-dimensional structure. This is in contrast to classical methods for which the computational expense grows exponentially with $d$, a major challenge for solution of high-dimensional PDEs \cite{bach2017breaking,bengio2000taking,han2018solving}. 

Understanding the performance of DNNs requires accounting for both optimal approximation error and optimization error. 
While one may prove existence of DNN parameters providing exponential convergence with respect to architecture size, in practice a number of issues conspire to prevent realizing such convergence. Several approximation theoretic works seek to understand the role of width/depth in the absence of optimization error \cite{he2018relu,daubechies2019nonlinear,yarotsky2017error,yarotsky2018optimal,opschoor2019deep}. In particular, Yarotsky and Opschoor et al. prove the existence of parameters for a deep neural network architecture that approximate algebraic operations, partitions of unity (POUs), and polynomials to exponential accuracy in the depth of the network.
This shows that sufficently deep DNNs may in theory learn a spectrally convergent $hp$-element space without a hand-tailored mesh by constructing a POU to localize polynomial approximation. In practice, however, such convergent approximations are not realized when training DNNs using gradient descent optimizers, even for smooth target functions \cite{fokina2019growing,adcock2020gap}.
The thesis of this work is to incorporate the POU and polynomial elements directly into a deep learning architecture. Rather than attempting to force a DNN to simultaneously perform localization and high-order approximation, introducing localized polynomial spaces frees the DNN to play to its strengths by focusing exclusively on partitioning space, as in classification problems.

An attractive property of the proposed architecture is its amenability to a fast training strategy. In previous work, we developed an optimizer which alternates between a gradient descent update of hidden layer parameters and a globally optimal least squares solve for a final linear layer \cite{cyr2019robust}; this was applied as well to classification problems \cite{patel2020block}. A similar strategy is applied in the current work: updating the POU with gradient descent before finding a globally optimal polynomial fit at each iteration ensures an optimal representation of data over the course of training.

While DNNs have \MG{been explored} as a means of solving high-dimensional PDEs \MG{\cite{geist2020numerical,han2018solving}}, optimization error \MG{prevents} a practical demonstration of convergence with respect to size of either data or model parameters \cite{beck2019full,wang2020understanding}. The relatively simple regression problem considered here provides a critical first example of how the optimization error barrier may be circumvented to provide accuracy competitive with finite element methods. 

\section{An abstract POU network}

Consider a \textit{partition of unity} $\Phi = \left\{\phi_\alpha(\mathbf{x})\right\}_{\alpha=1}^{N_\text{part}}$ satisfying $\sum_\alpha \phi_\alpha(\mathbf{x}) = 1$ and $\phi_\alpha(\mathbf{x}) \ge 0$ for all $\mathbf{x}$. We work with the approximant
\begin{equation}\label{eq:pou}
y_{\text{POU}}(\mathbf{x}) = \sum_{\alpha=1}^{N_\text{part}} \phi_\alpha(\mathbf{x}) \sum_{\beta=1}^{\text{dim}(V)} c_{\alpha,\beta} P_\beta(\mathbf{x}),
\end{equation}
where $V = \text{span}\left\{P_\beta\right\}$. 
For this work, we take $V$ to be the space  $\pi_m(\mathbb{R}^d)$ of polynomials of order $m$\MG{, while $\Phi$ is parametrized as a neural network with weights and biases $\bm{\xi}$ and output dimension $N_{\text{part}}$}:
\begin{equation}\label{eq:DNN_formula}
\MG{\phi_\alpha(\mathbf{x}; \bm{\xi}) = \big[  \mathcal{NN}(\mathbf{x}; \bm{\xi})  \big]_{\alpha},
\quad
1 \le \alpha \le N_{\text{part}}}.
\end{equation}
\MG{We consider two architectures for $\mathcal{NN(\mathbf{x};\bm{\xi})}$ to be specified later.}
Approximants of the form \eqref{eq:pou} allow a ``soft'' localization of the basis \MG{elements $P_{\beta}$} to \MG{an implicit} partition of space \MG{parametrized by the $\phi_\alpha$}. While approximation with broken polynomial spaces corresponds to taking $\Phi$ to consist of characteristic functions on the cells of a computational mesh, the parametrization of $\Phi$ by a DNN generalizes more broadly to differentiable partitions of space. \MG{For applications of partitions of unity in numerical analysis, see \citet{strouboulis2001generalized,fries2010extended,wendland2002fast};
for their use in differential geometry, see
\citet{spivak2018calculus,hebey2000nonlinear}}. 

In a traditional numerical procedure, $\Phi$ is constructed prior to fitting $c_{\alpha,\beta}$ to data through a geometric ``meshing'' process. We instead work with a POU $\Phi^{\bm{\xi}}$ \MG{in the form of a DNN \eqref{eq:DNN_formula} in which the} weights and biases $\bm{\xi}$, which are trained to fit the data. We therefore fit both the localized basis coefficients $\bm{c} = [c_{\alpha,\beta}]$ and the localization itself simultaneously by solving the optimization problem
\begin{equation}\label{eq:loss}
\underset{{\bm{\xi}},\mathbf{c}}{\text{argmin}} \underset{i\in\mathcal{D}}{\sum} \left| \sum_{\alpha=1}^{N_\text{part}} \phi_\alpha(\mathbf{x}_i,{\bm{\xi}}) \sum_{\beta=1}^{\text{dim}(V)} c_{\alpha,\beta} P_\beta(\mathbf{x}_i) - {y}_i \right|^2.
\end{equation}

\section{Error analysis and architecture design}
Before specifying a choice of \MG{architecture} for $\Phi^{\bm{\xi}}$ \MG{in \eqref{eq:DNN_formula}}, we present a basic estimate of the optimal training error using the POU-Net architecture to highlight desirable properties for the partition to have. 
We denote by $\text{diam}(A)$ the diameter of a set $A \subset{\mathbb{R}^d}$.

\theorem{Consider an approximant $y_{\text{POU}}$ of the form \eqref{eq:pou} with $V = \pi_m(\mathbb{R}^d)$. If $y(\cdot) \in C^{m+1}(\Omega)$ and  $\bm{\xi}^*, \bm{c}^*$ solve \eqref{eq:loss} to yield the approximant ${y}^*_{\text{POU}}$, then
\begin{equation}\label{eq:training_error_estimate}
\|{y}^*_{\text{POU}} - {y} \|_{\ell_2(\mathcal{D})}^2 \leq C_{m,y}\, \underset{\alpha} {\text{max}}\, \text{diam}\left(\text{supp}(\phi^{\bm{\xi}}_\alpha)\right)^{m+1}
\end{equation}
where $\|{y}^*_{\text{POU}} - {y} \|_{\ell_2(\mathcal{D})}$ denotes the root-mean-square norm over the training data pairs in $\mathcal{D}$,
\begin{equation}
\MG{ \|{y}^*_{\text{POU}} - {y} \|_{\ell_2(\mathcal{D})} = 
\sqrt{\frac{1}{N_{\text{data}}}
\sum_{(\mathbf{x},y) \in \mathcal{D}} 
\left({y}^*_{\text{POU}}(\mathbf{x}) - {y}(\mathbf{x})\right)^2
}
,}
\end{equation}
\MG{and}
\begin{equation}
\MG{C_{m,y} = \| y \|_{C^{m+1}(\Omega)}.}
\end{equation}\label{thm:training_error}}
\proof{
For each $\alpha$, take $q_\alpha \in \pi_m(\mathbb{R}^d)$ to be the $m$th order Taylor polynomial of $y(\cdot)$ centered at any point of $\text{supp}(\phi^{\bm{\xi}}_\alpha)$. Then for all $\mathbf{x}\in \text{supp}(\phi^{\bm{\xi}}_\alpha)$, 
\begin{equation}\label{eq:taylor_error}
|q_\alpha(\mathbf{x})-y(\mathbf{x})| \leq C_{m,y} \, \text{diam}\left(\text{supp}(\phi^{\bm{\xi}}_\alpha)\right)^{m+1}.
\end{equation}
Define the approximant
$\widetilde{{y}}_{\text{POU}} = \sum_{\alpha=1}^{N_\text{part}} \phi^{\bm{\xi}}_\alpha(\mathbf{x}) q_\alpha(\mathbf{x})$, which is of the form \eqref{eq:pou} and represented by feasible $(\bm{\xi},\bm{c})$.
Then by definition of $y^*_{\text{POU}}$ and \eqref{eq:loss}, we have
\begin{align}
\|{y}^*_{\text{POU}}(\mathbf{x}) - &{y}(\mathbf{x}) \|_{\ell_2(\mathcal{D})}^2 
\le
\|\widetilde{y}_{\text{POU}}(\mathbf{x}) - {y}(\mathbf{x}) \|_{\ell_2(\mathcal{D})}^2  \\
&=
\left\|
\sum_{\alpha=1}^{N_\text{part}} \phi^{\bm{\xi}}_\alpha(\mathbf{x}) q_\alpha(\mathbf{x})
- {y}(\mathbf{x}) \sum_{\alpha=1}^{N_\text{part}} \phi^{\bm{\xi}}_\alpha(\mathbf{x})
\right\|_{\ell_2(\mathcal{D})}^2 \\
&=
\left\|
\sum_{\alpha=1}^{N_\text{part}} \phi^{\bm{\xi}}_\alpha(\mathbf{x}) 
\left(q_\alpha(\mathbf{x}) - {y}(\mathbf{x}) \right)
\right\|_{\ell_2(\mathcal{D})}^2.
\end{align}
For each $\mathbf{x} = \mathbf{x}_i \in \mathcal{D}$, if $\mathbf{x} \in \text{supp}(\mathcal{D})$, then we apply \eqref{eq:taylor_error}; otherwise, the summand $\phi^{\bm{\xi}}_\alpha(\mathbf{x}) 
\left(q_\alpha(\mathbf{x}) - {y}(\mathbf{x}) \right)$ vanishes. So
\begin{align}
\|&{y}^*_{\text{POU}}(\mathbf{x}) - {y}(\mathbf{x}) \|_{\ell_2(\mathcal{D})}^2 \\
&\le
\left\|
\sum_{\alpha=1}^{N_\text{part}}
C_{m,y} \ \text{diam}\left(\text{supp}(\phi^{\bm{\xi}}_\alpha)\right)^{m+1}
\phi^{\bm{\xi}}_\alpha(\mathbf{x}) 
\right\|_{\ell_2(\mathcal{D})}^2 \\
&\le
C_{m,y} \ \underset{\alpha}{\text{max}} \ \text{diam}\left(\text{supp}(\phi^{\bm{\xi}}_\alpha)\right)^{m+1}
\left\|
\sum_{\alpha=1}^{N_\text{part}}
\phi^{\bm{\xi}}_\alpha(\mathbf{x}) 
\right\|_{\ell_2(\mathcal{D})}^2 \\
&\le
C_{m,y} \ \underset{\alpha}{\text{max}} \ \text{diam}\left(\text{supp}(\phi^{\bm{\xi}}_\alpha)\right)^{m+1}.
\end{align}}

\MG{Theorem \ref{thm:training_error}} indicates that \MG{for smooth $y$, the resulting convergence rate \MG{of the training error is independent of the specific choice of parameterization of the $\phi_{\alpha}^{\bm{\xi}}$}, and depends only upon \MG{their} support.} Further, the error does not explicitly depend upon the dimension of the problem; if \MG{the trained} $\Phi^{\bm{\xi}}$ \MG{encodes a covering by supp$(\phi_\alpha^{\bm{\xi}})$} of a low dimensional manifold \MG{containing the data locations $\mathbf{x}$} with latent dimension $d_{\mathcal{M}} \ll d$, then the approximation \MG{cost} scales only with $\text{dim}(V)$ and thus may break the curse of dimensionality, e.g. for linear polynomial approximation $\text{dim}(V)=d+1.$ If the parameterization is able to find compactly supported quasi-uniform partitions \MG{such that the maximum diameter in \eqref{eq:training_error_estimate} scales as $N_{\text{part}}^{-1/{d_{\mathcal{M}}}}$,
the training loss will exhibit a scaling of $N_{\MG{\text{part}}}^{-({m+1})/{d_\mathcal{M}}}$}.

\MG{The above analysis indicates the advantage of enforcing highly localized, compact support of the POU functions. However, in practice we found that for a POU parametrized by a shallow RBF-Net networks (described below), rapidly decaying but globally supported POU functions exhibited more consistent training results. This is likely due to a higher tolerance to initializations poorly placed with respect to the data locations. Similarly, when the POU is parametrized by a deep ResNet, we obtained good results using ReLU activation functions while training with a regularizer to promote localization (Algorithm \ref{thealg2}). Thus, the properties playing a role in the above analysis -- localization of the partition functions and distribution of their support over a latent manifold containing the data -- are the motivation for the architectures and training strategies we consider below.} \\

\noindent \textbf{POU \#1 - RBF-Net:} A shallow RBF-network \cite{broomhead1988radial,billings1995radial} implementation of $\Phi_{\bm{\xi}}$ is given by 
\eqref{eq:pou} and
\begin{equation}
\MG{
\phi_\alpha = \frac{\exp \left(-|x-{\bm{\xi}}_{1,\alpha}|^2/{\bm{\xi}}_{2,\alpha}^2\right)}{\sum_\beta\exp \left(-|x-{\bm{\xi}}_{1,\beta}|^2/{\bm{\xi}}_{2,\beta}^2\right)}.}
\end{equation}
Here, $\bm{\xi}_1$ denotes the RBF centers and $\bm{\xi}_2$ denotes RBF shape parameters, both of which evolve during training. \MG{A measure of the localization of these functions can be taken to be} the magnitude of $\bm{\xi}_1$. \MG{Such an architecture works well for approximation of smooth functions, but} the $C_\infty$ continuity of $\Phi_{\bm{\xi}}$ \MG{causes difficulty in the approximation} of piecewise smooth functions. \\

\noindent \textbf{POU \#2 - ResNet:} We compose a residual network architecture \cite{he2016deep} with a softmax \MG{layer} $\mathcal{S}$ \MG{to define \eqref{eq:DNN_formula}.} For the experiments considered we use a ReLU activation, allowing representation of functions with with discontinuities in their first derivative. \\

\noindent \textbf{Initialization:} All numerical experiments take data supported within the unit hypercube $\Omega\subset[0,1]^d$. To initialize the \MG{POU \#1} architecture we use unit shape parameter and uniformly distribute centers $\mathbf{x}_1 \sim \mathcal{U}([0,1]^d)$. \MG{We initialize POU \#2 with the Box} initialization \cite{cyr2019robust}. \MG{We found that these initializations} provide an initial set of partitions \MG{sufficiently }``well-distributed'' throughout $\Omega$ \MG{ for successful training}.

\section{Fast optimizer}\label{sec:fast_optimizer}

The least-squares structure of \eqref{eq:loss} allows application of the least-squares gradient descent (LSGD) block coordinate descent strategy \cite{cyr2019robust}. At each epoch, one may hold the hidden parameters ${\bm{\xi}}$ fixed to obtain a least squares problem for the optimal polynomial coefficients $\mathbf{c}$. The step may be concluded by then taking a step of gradient descent for the POU parameters, therefore evolving the partitions along the manifold corresponding to optimal representation of data; for details see \cite{cyr2019robust}.

Algorithm \ref{thealg1} with $\lambda=0$ specifies the application of LSGD to Eqn. \eqref{eq:loss}. We will demonstrate that while effective, several of the learned partition \MG{functions $\phi_\alpha$} may ``collapse'' to near-zero values \MG{everywhere}. To remedy this, we will also consider a pre-training step in Algorithm \ref{thealg2}, which adds an $\ell_2$ regularizer to the polynomial coefficients. The intuition behind this is that a given partition \MG{regresses data using an element of the form}
$c_{\alpha,\beta} \MG{\phi_\alpha}P_\beta$. 
If $\phi_\alpha$ is scaled by a small $\delta>0$, the LSGD solver may pick up a scaling \MG{${1}/{\delta}$} for $c_{\alpha,\beta}$ and achieve the same approximation. Limiting the coefficients thus indirectly penalizes this mode of partition \MG{function} collapse, promoting more quasi-uniform partitions of space. 

\RestyleAlgo{plainruled}
\begin{algorithm}[t]
 \KwData{${\bm{\xi}}_{\text{old}},\mathbf{c}_{\text{old}}$}
 \KwResult{${\bm{\xi}}_{\text{new}}, \mathbf{c}_{\text{new}}$}
 \SetKwFunction{FMain}{LSGD}
 \SetKwProg{Fn}{Function}{:}{}
  
  \Fn{\FMain{
  ${\bm{\xi}}_{\textup{old}}$,
  $\mathbf{c}_{\textup{old}}$,
  $n_{\textup{epoch}}$,
  $\lambda$,
  $\rho$,
  $n_{\textup{stag}}$}} 
  {
    ${\bm{\xi}},\mathbf{c} \gets {\bm{\xi}}_{\text{old}},\mathbf{c}_{\text{old}}$\;
    \For{$i \in \{1,...,n_{\textup{epoch}}\}$}
    {
        $\mathbf{c} \gets \text{LS}({\bm{\xi}},\lambda)$ \tcp*[l]{solve Eqn.~\eqref{eq:pou} with a regularizer 
        $\lambda \| \mathbf{c} \|_{F}^2$}
        ${\bm{\xi}} \gets \text{GD}(\mathbf{c})$\;
        \If{\texttt{LSGD} \text{stagnates more than} $n_{\textup{stag}}$} { 
        $\lambda \gets \rho \lambda $
     }
  }
  $\mathbf{c}_{\text{new}} \gets \text{LS}({\bm{\xi}},0)$\;
  ${\bm{\xi}}_{\text{new}} \gets {\bm{\xi}}$\;
  }
\caption{The regularized least-squares gradient descent (LSGD) method. Setting $\lambda=0$ recovers the original LSGD method \cite{cyr2019robust}.}
 \label{thealg1}
\end{algorithm}

\RestyleAlgo{plainruled}
\begin{algorithm}[t]
 \KwData{${\bm{\xi}}_{\text{old}},\mathbf{c}_{\text{old}}$}
 \KwResult{${\bm{\xi}}_{\text{new}}, \mathbf{c}_{\text{new}}$}
 \SetKwFunction{FMain}{Two-phase-LSGD}
 \SetKwProg{Fn}{Function}{:}{}
  
  \Fn{\FMain{
  ${\bm{\xi}}_{\textup{old}}$,
  $\mathbf{c}_{\textup{old}}$,
  $n_{\textup{epoch}},n_{\textup{epoch}}^{\textup{pre}}$,
  $\lambda$,
  $\rho$,$n_{\textup{stag}}$}} {
      ${\bm{\xi}},\mathbf{c} \gets
       \texttt{LSGD}({\bm{\xi}}_{\textup{old}},\mathbf{c}_{\textup{old}},n_{\text{epoch}}^{\text{pre}},\lambda,\rho,n_{\textup{stag}})$ \tcp*[l]{Phase 1: LSGD with a regularizer}

      ${\bm{\xi}}_{\textup{new}},\mathbf{c}_{\textup{new}} \gets \texttt{LSGD}({\bm{\xi}},\mathbf{c},n_{\text{epoch}},0,0,n_{\text{epoch}})$ 
      \tcp*[l]{Phase 2: LSGD without a regularizer}
 
 }
 \caption{The two-phase LSGD method with the $\ell^2$-regularized least-squares solves.}
 \label{thealg2}
\end{algorithm}

\section{Numerical experiments}
In this section, we assess the performance of POUnets with \MG{the} two POU architectures. We implement the proposed algorithms in \textsc{Python}, construct POUnets using \textsc{Tensorflow 2.3.0} \cite{abadi2016tensorflow}, and employ \textsc{NumPy} and \textsc{Scipy} packages for generating training data and solving the least squares problem. For all considered neural networks, training is performed by batch gradient descent using the Adam optimizer \cite{kingma2014adam}. For a set of $d$-variate polynomials $\{P_{\beta}\}$, we choose truncated Taylor polynomials of maximal degree $m_{\max}$, providing to $\frac{(m_{\max}+d)!}{(m_{\max}!)(d!)}$ polynomial coefficients per partition. The coefficients are initialized via sampling from the unit normal distribution.

\subsection{\MG{Smooth functions}}
We consider an analytic function as our first benchmark, specifically the sine function defined on a cross-shaped \MG{one}-dimensional manifold embedded in 2-dimensional domain $[-1,1]^2$ 
\begin{equation*}
    \mathbf{y}(\mathbf{x}) = \left\{ 
    \begin{array}{cc}
    \sin (2\pi x_1), & \text{if }x_2 = 0,\\
    \sin (2\pi x_2), & \text{if }x_1 = 0.
    \end{array}
    \right.
\end{equation*}

\begin{figure}[!b]
    \centering
    \subfigure[POUnets]{\includegraphics[scale=.4]{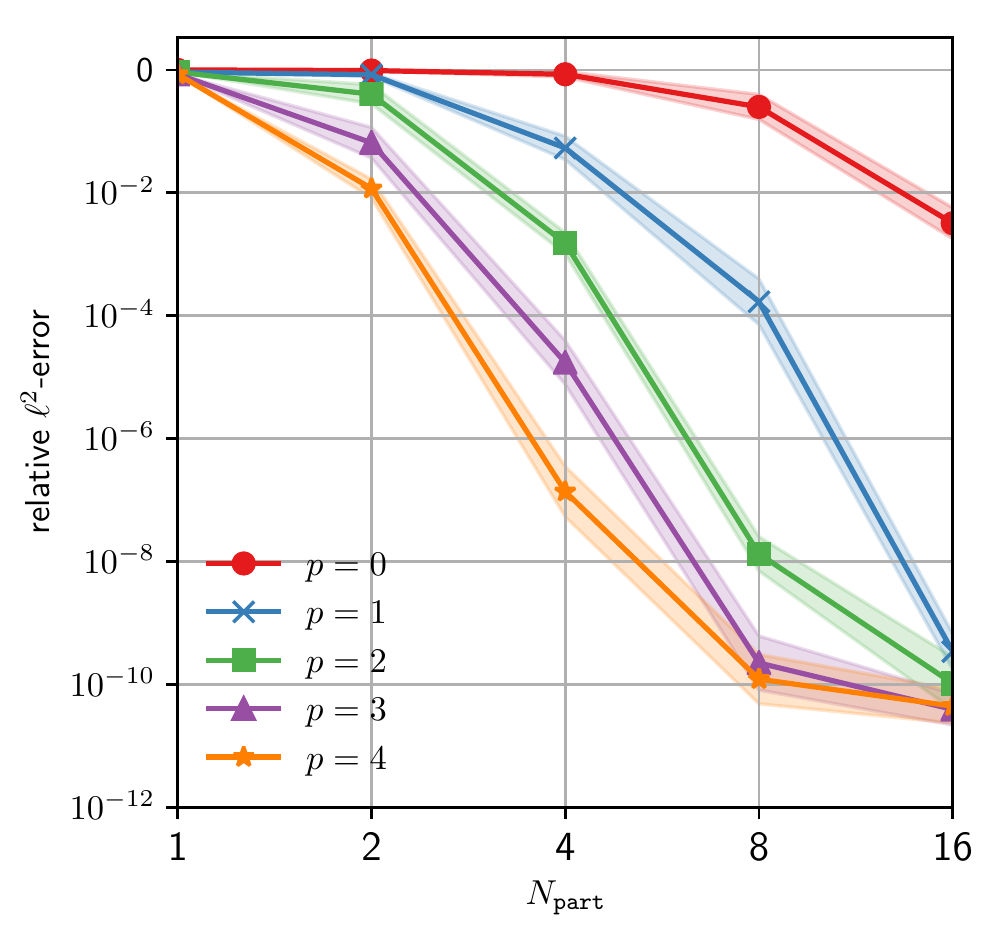}\label{fig:rbf_sub}}\hspace{1mm}
    \subfigure[MLPs]{\includegraphics[scale=.4]{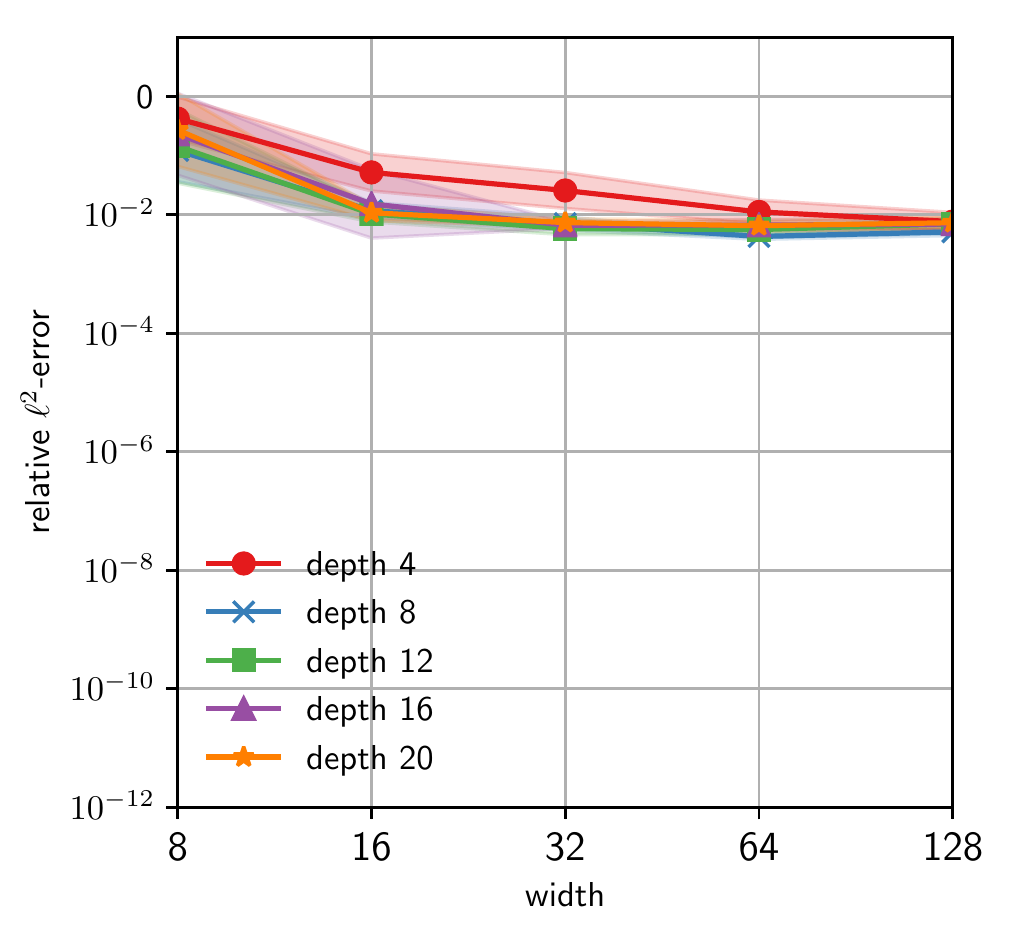}}
    \caption{Relative $\ell^2$-errors \MG{(log-log scale)} of approximants produced by POUnets \MG{with RBF-Net partition functions} for varying $N_{\text{part}}$ and varying $m_{\max}$ (left) and standard MLPs for varying width and depth (right).} 
    \label{fig:rbf_regression_conv}
\end{figure}

We test \MG{RBF-Nets} for varying number of partitions, $N_{\text{part}} = \{1,2,4,8,16\}$ and the maximal polynomial degrees $\{0,1,2,3,4\}$. For training, we collect data $x_i,i=1,2$ by uniformly sampling from the domain [-1,1] for 501 samples, i.e., in total, 1001 $\{((x_1,x_2), \mathbf{y}(\mathbf{x})\}$-pairs \MG{after removing the duplicate point on the} origin. We initialize centers \MG{of the RBF basis functions} by sampling \MG{uniformly} from the domain $[-1,1]^2$ and initialize shape parameters as ones. We then train \MG{RBF-Nets} by using the LSGD method with $\lambda=0$ (Algorithm~\ref{thealg1}) with the initial learning rate for Adam set to $10^{-3}$. The maximum number of epochs $n_{\text{epoch}}$ is set as 100, and we choose the centers and shape parameters that yield the best result on the training loss during the specified epochs.

Figure~\ref{fig:rbf_sub} reports the relative $\ell^2$-errors of approximants produced by POUnets for varying $N_{\text{part}}$ and varying $p$. The results are obtained from 10 independent runs for each value of $N_{\text{part}}$ and $p$, with a single log-normal standard deviation. Algebraic convergence is observed of order increasing with polynomial degree before saturating at $~10^{-10}$. \MG{We} conjecture this is due to eventual loss of precision due to the ill-conditioning of the least squares matrix\MG{;} however\MG{,} we leave a formal study and treatment for a future work.

In comparison to the performance achieved by POUnets, we assess the performance of the standard MLPs with varying depth \{4,8,12,16,20\} and width \{8,16,32,64,128\}. The standard MLPs are trained by using Adam with an initial learning rate $10^{-3}$ and the maximum number of epochs 1000. As opposed to the convergence behavior of RBF-Net-based POUnets, the standard MLP briefly exhibits roughly first order convergence before stagnating around an error of $~10^{-2}$.

\begin{figure}[!b]
    \centering
    \subfigure[Triangular waves]{\includegraphics[scale=.5]{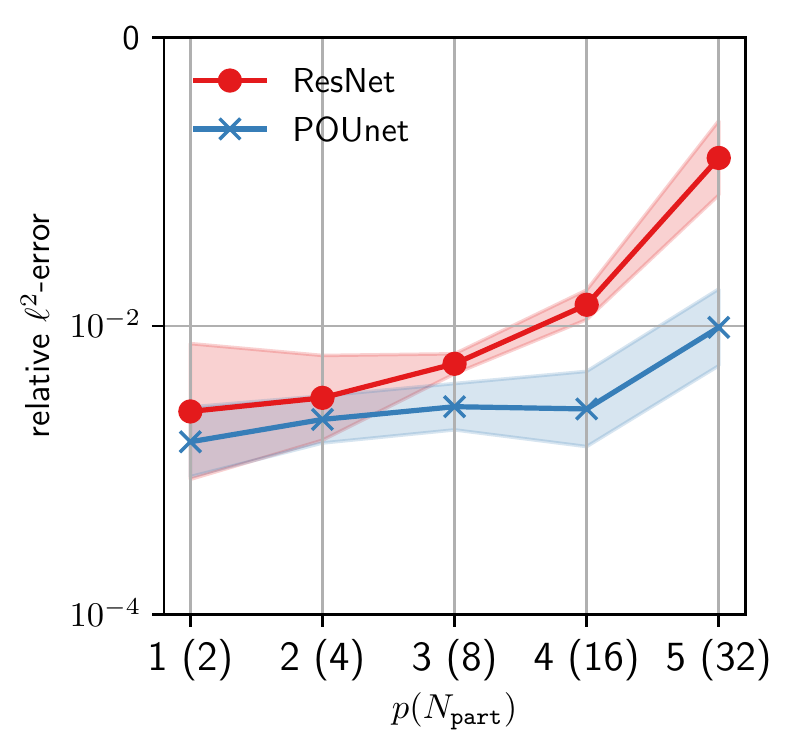}}
    \subfigure[Quadratic waves]{\includegraphics[scale=.5]{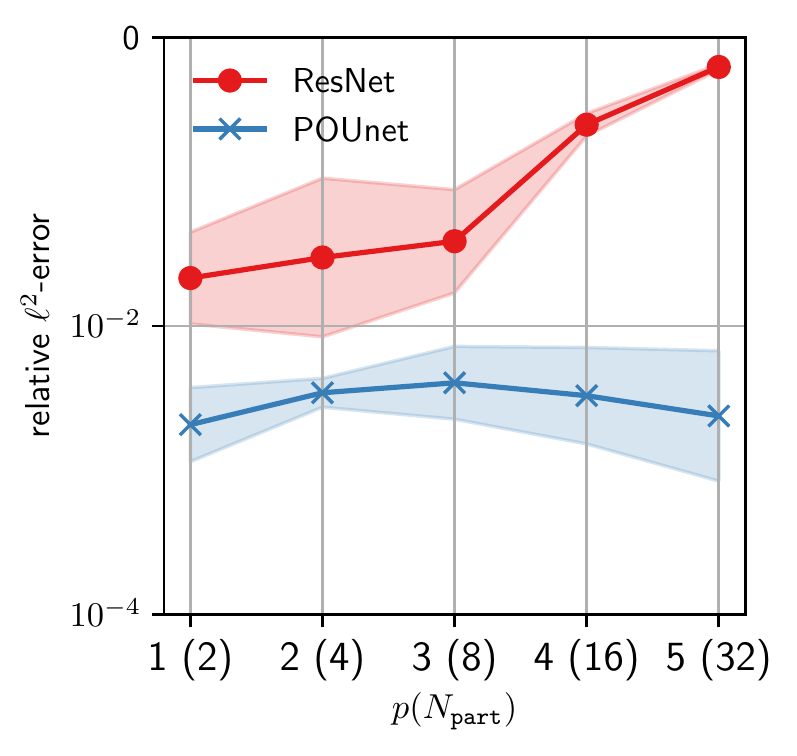}}
    \caption{Relative $\ell^2$-errors \MG{(log-log scale)} of approximants produced by the standard ResNets and POUnets \MG{with ResNet partition functions} for varying $N_{\text{part}}$ for approximating the target functions with varying $p$.} 
    \label{fig:regression_errors}
\end{figure}

\subsection{\MG{Piecewise Smooth Functions}}
We next consider piecewise linear and piecewise quadratic functions: triangle waves with varying frequencies, i.e., $\mathbf{y}(x) = \textsc{TRI}(x;p)$, and their quadratic variants $\mathbf{y}(x) = \textsc{TRI}^2(x;p)$, where 
\begin{equation}
    \textsc{TRI}(x;p) = 2 \left \vert p x - \left \lfloor px + \frac{1}{2} \right \rfloor \right\vert -1.
\end{equation}
We study the introduction of increasingly many discontinuities by increasing the frequency $p$. Reproduction of \MG{such} sawtooth functions by ReLU networks via both \MG{wide networks} \cite{he2018relu} and \MG{very deep networks \cite{telgarsky2016benefits} can be} achieved theoretically via construction of carefully chosen \MG{architecture and} weights/biases, but to our knowledge has not been achieved via \MG{standard} training.

The \MG{smoothness} possessed by the \MG{RBF-Net partition functions (POU \#1)} precludes \MG{rapidly convergent approximation of piecewise smooth functions}, and we instead employ ResNets \MG{(POU \#2)} for $\Phi^{\bm{\xi}}$ \MG{in this case, as the piecewise linear nature of such partition functions is better suited}. As a baseline for the performance comparison, we apply regression of the same data over a mean square error using standard ResNets,  $y_{\text{MLP}}(x)$. The standard ResNets share the same architectural design and hyperparameters with the \MG{ResNets used to parametrize the POU functions in the POUnets}. The only exception is the output layer; the standard ResNets produce a scalar-valued output, whereas the \MG{POU parametrization produce}s a vector-valued output, which is followed by the softmax activation. 

\begin{figure}[!t]
    \centering
    \includegraphics[scale=.65]{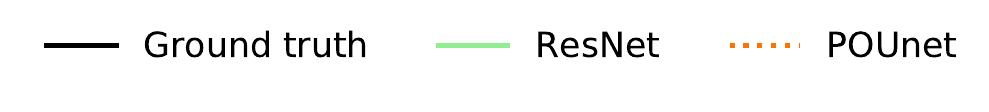}\vspace{-5mm}\\
    \subfigure[$N_\text{part}=16$]{\includegraphics[scale=.55]{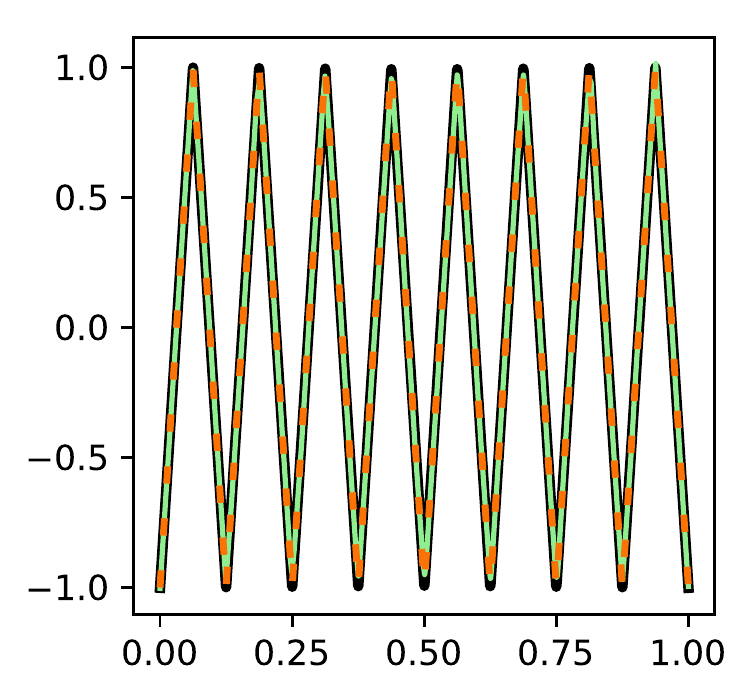}}
    \subfigure[$N_\text{part}=16$]{\includegraphics[scale=.55]{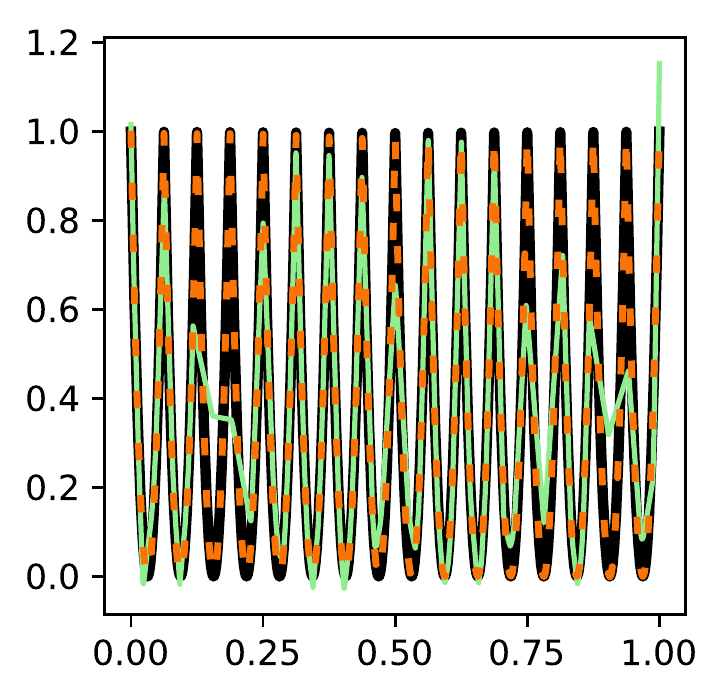}}\\
    \subfigure[$N_\text{part}=32$]{\includegraphics[scale=.55]{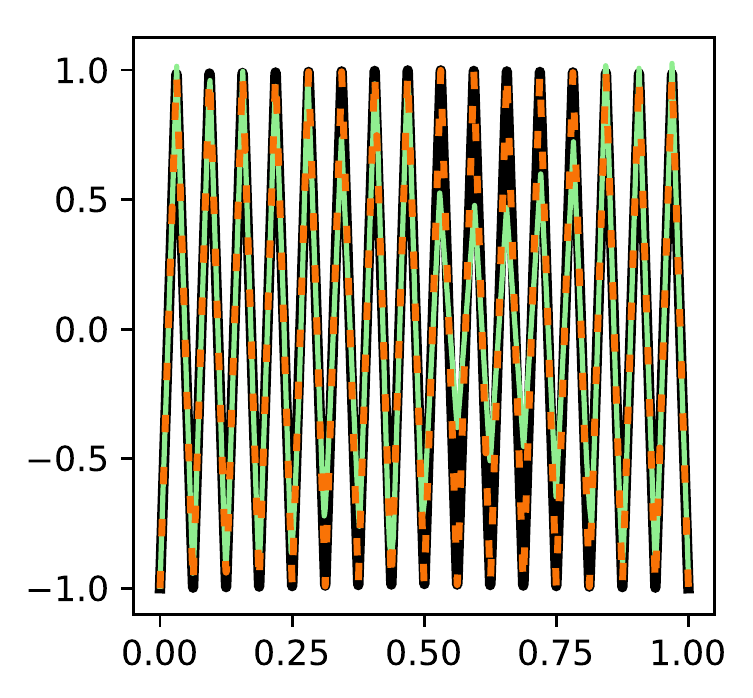}\label{fig:regression_snapshots_sub3}}
    \subfigure[$N_\text{part}=32$]{\includegraphics[scale=.55]{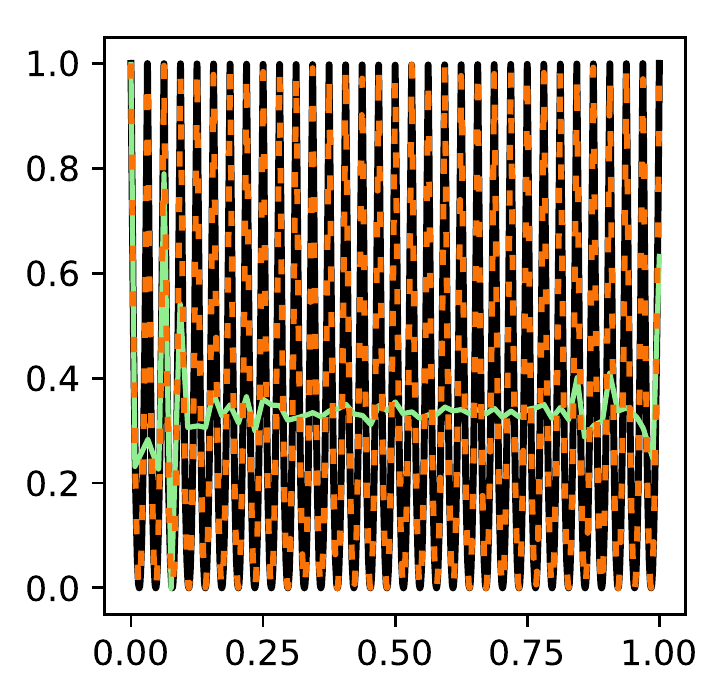}\label{fig:regression_snapshots_sub4}}\\
    \hspace{1mm}
    \subfigure[$N_\text{part}=32$ (zoomed-in)]{\includegraphics[scale=.51]{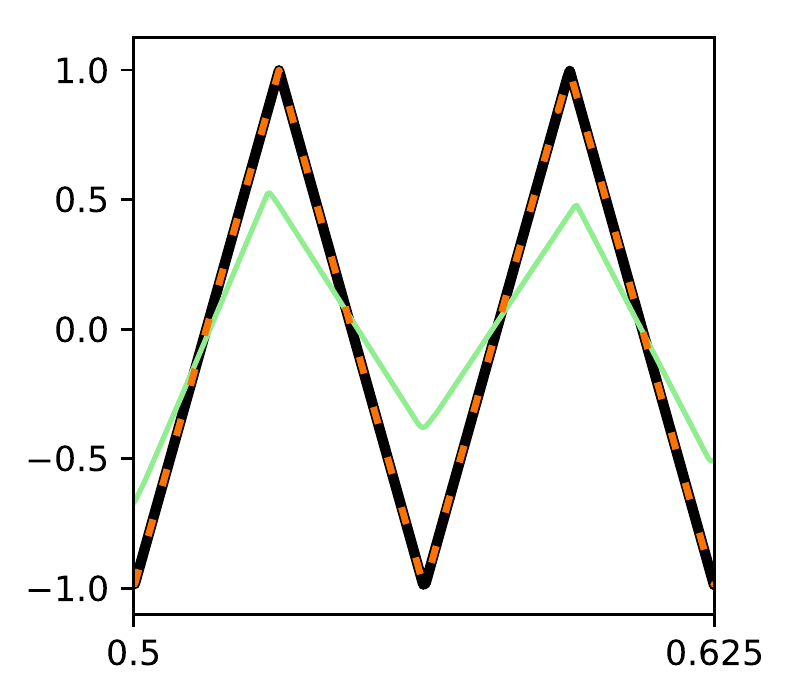}\label{fig:regression_snapshots_sub5}}
    \subfigure[$N_\text{part}=32$ (zoomed-in)]{\includegraphics[scale=.51]{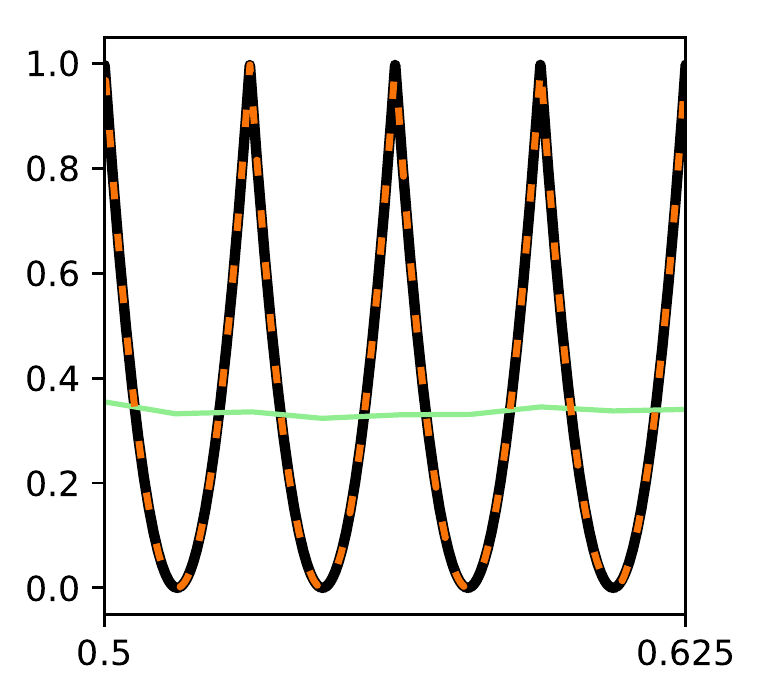}\label{fig:regression_snapshots_sub6}}
    \caption{Snapshots of target functions $\mathbf{y}(\mathbf{x})$ and approximants produced by ResNet and POUnet (i.e., $y_{\text{POU}}(\mathbf{x})$) are depicted in \MG{black, light green, and orange,} respectively. The target function correspond to triangular waves (left) and their quadratic variants (right). The bottom row depicts the snapshots in the domain [0.5,0.625].} 
    \label{fig:regression_snapshots}
\end{figure}

We consider five frequency parameters for both target functions, $p=\{1,2,3,4,5\}$, which results in piecewise linear and quadratic functions with $2^p$ pieces. Based on the number of pieces in the target function, we scale the width of the baseline neural networks and POUnets as $4\times 2^p$, while fixing the depth as 8, and for POUnets the number of partitions are set as $N_{\text{part}}=2^p$. For POUnets, we choose the maximal degree of polynomials to be $m_{\max}=1$ and $m_{\max}=2$ for the piecewise linear and quadratic target functions, respectively.

\begin{figure*}[t]
    \subfigure[Phase 1 (0)]{\includegraphics[height=28.5mm,width=28.5mm]{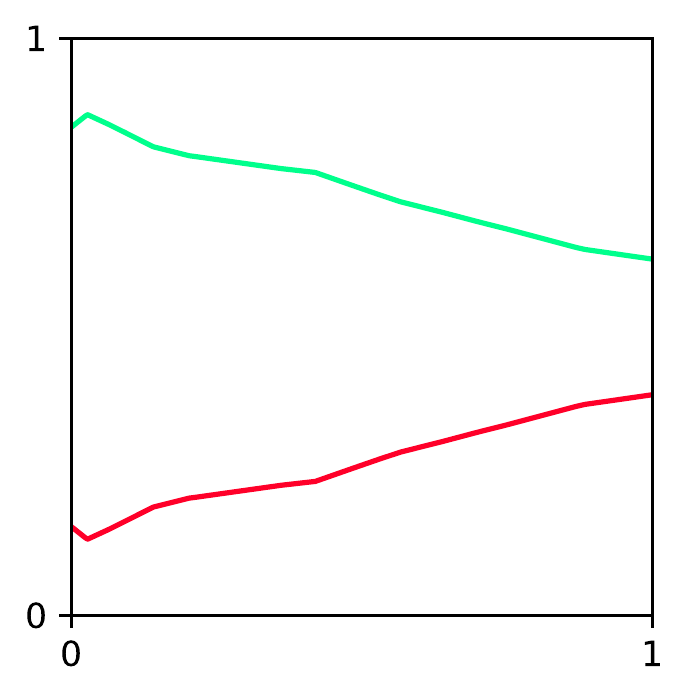}\label{fig:tri0_sub1}}
    \subfigure[Phase 1 (15)]{\includegraphics[height=28.5mm,width=28.5mm]{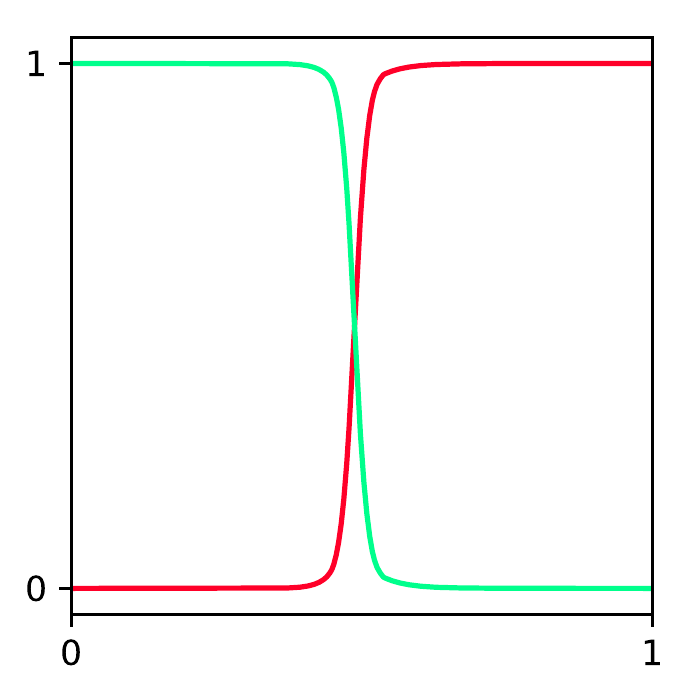}\label{fig:tri0_sub2}}
    \subfigure[Phase 1 (30)]{\includegraphics[height=28.5mm,width=28.5mm]{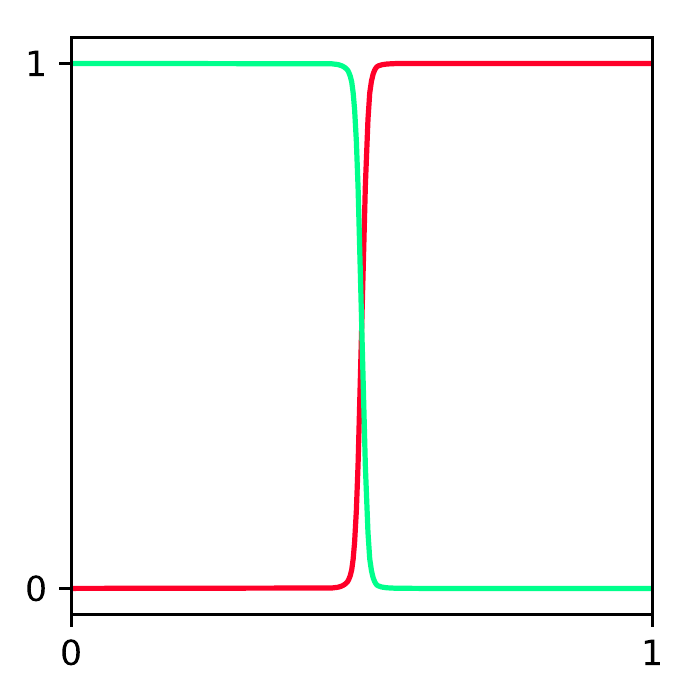}\label{fig:tri0_sub3}}
    \subfigure[Phase 1 (60)]{\includegraphics[height=28.5mm,width=28.5mm]{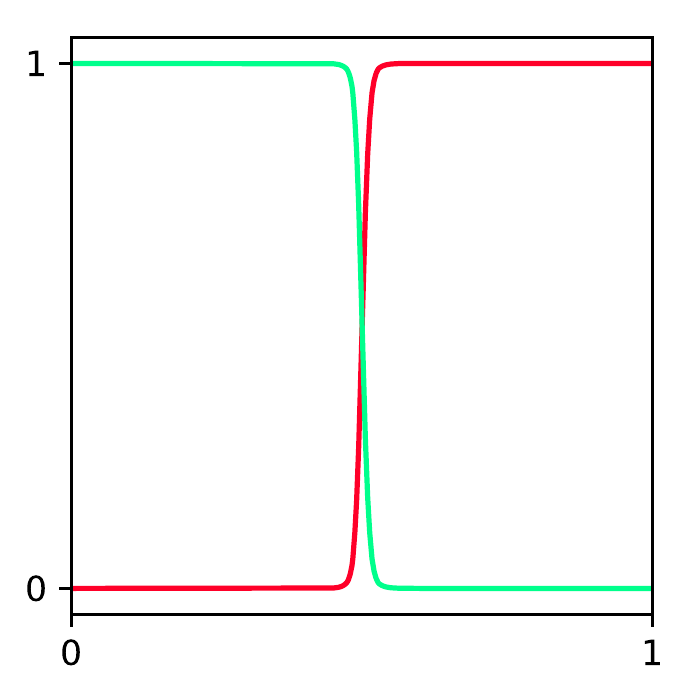}\label{fig:tri0_sub4}}
    \subfigure[Phase 2 (1000)]{
    \includegraphics[height=28.5mm,width=28.5mm]{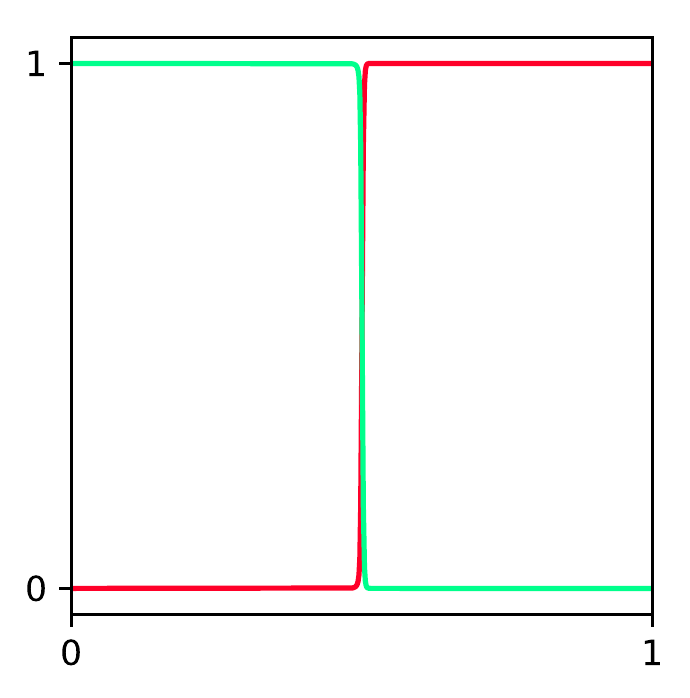}\label{fig:tri0_sub5}}
    \subfigure[Approximation]{\includegraphics[height=28.5mm,width=28.5mm]{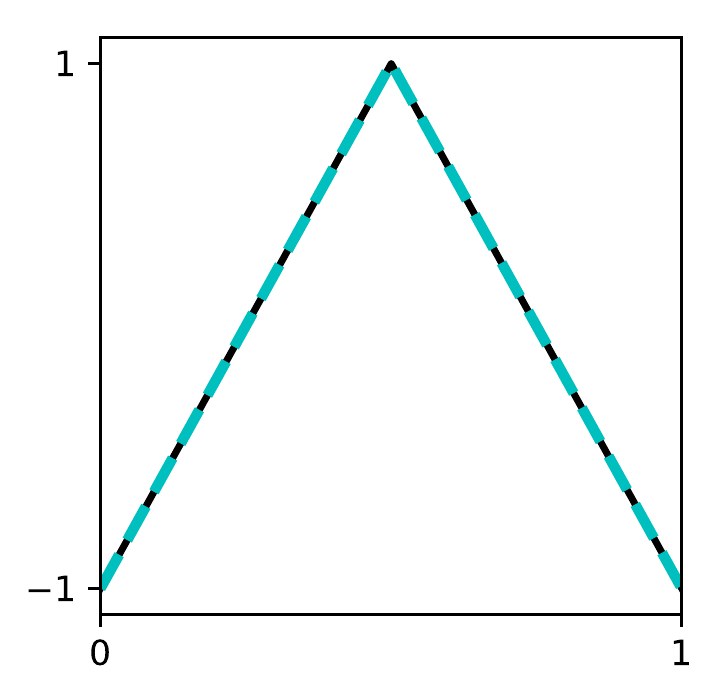}\label{fig:tri0_sub6}}\\
    \subfigure[Phase 1 (0)]{\includegraphics[height=28.5mm,width=28.5mm]{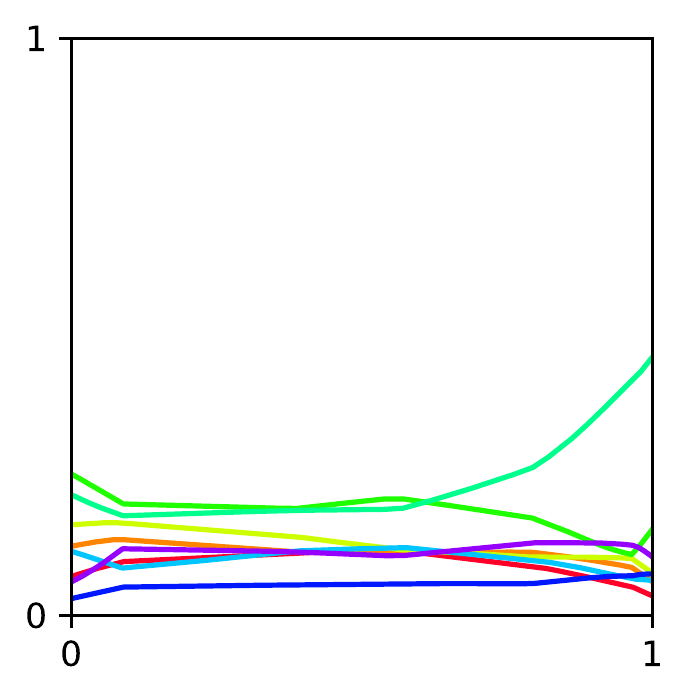}\label{fig:tri2_sub1}}
    \subfigure[Phase 1 (30000)]{\includegraphics[height=28.5mm,width=28.5mm]{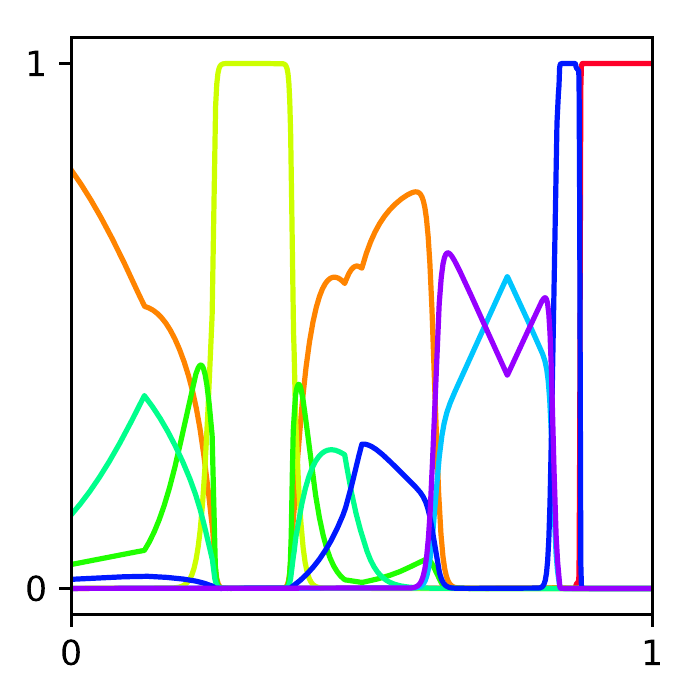}\label{fig:tri2_sub2}}
    \subfigure[Phase 1 (60000)]{\includegraphics[height=28.5mm,width=28.5mm]{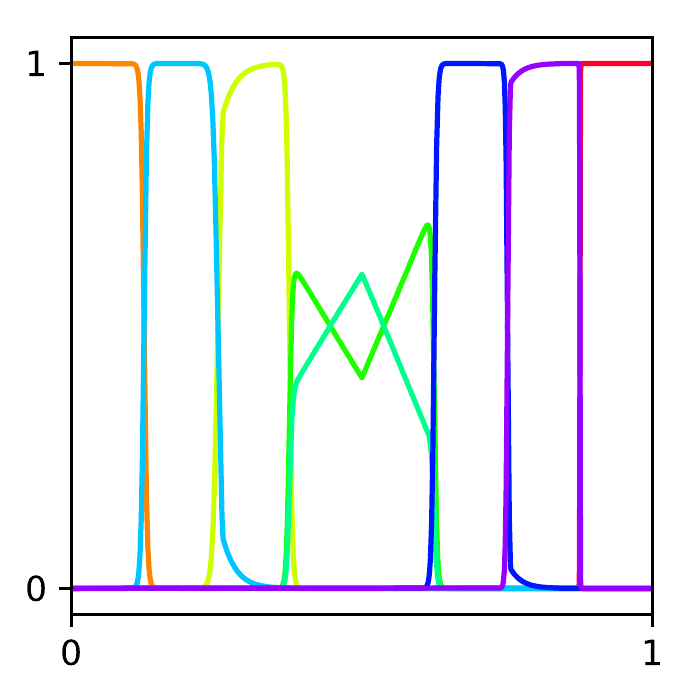}\label{fig:tri2_sub3}}
    \subfigure[Phase 1 (90000)]{\includegraphics[height=28.5mm,width=28.5mm]{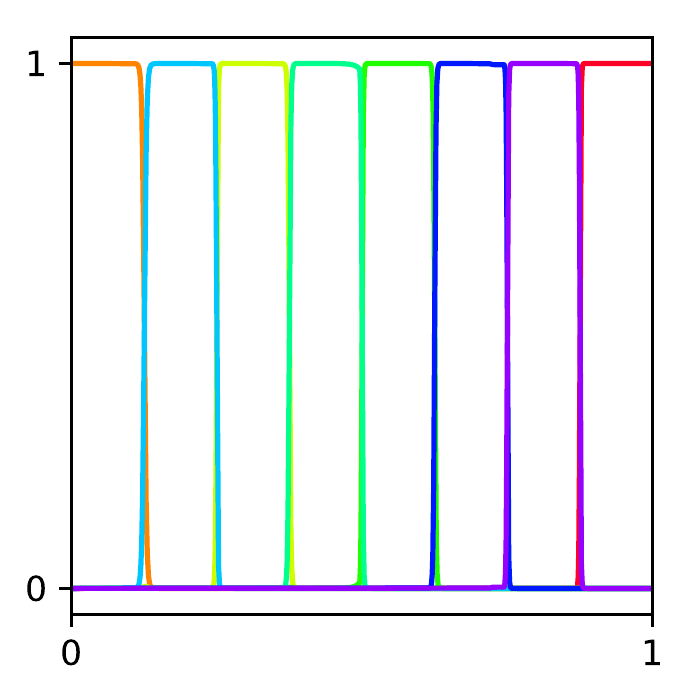}\label{fig:tri2_sub4}}
    \subfigure[Phase 2 (1000)]{
    \includegraphics[height=28.5mm,width=28.5mm]{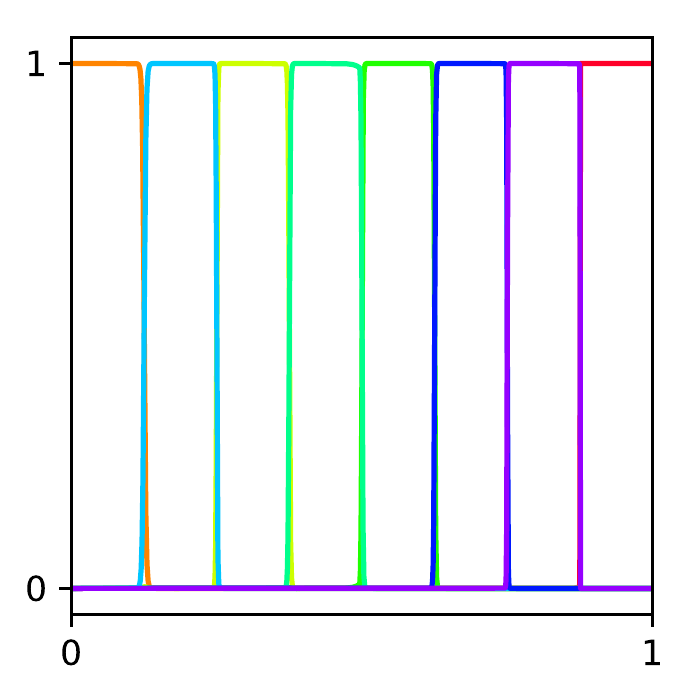}\label{fig:tri2_sub5}}
    \subfigure[Approximation]{\includegraphics[height=28.5mm,width=28.5mm]{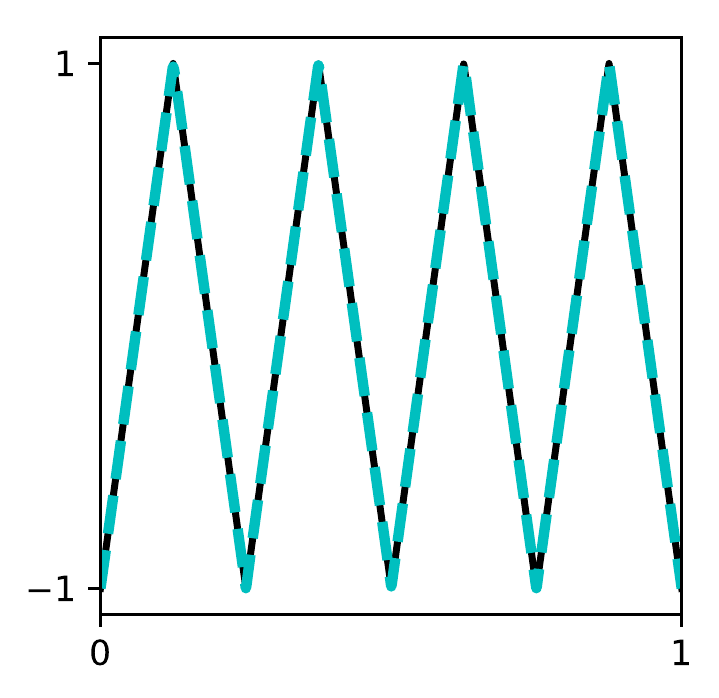}\label{fig:tri2_sub6}}
    \caption{Triangular wave with two pieces (top) and triangular wave with eight pieces (bottom): Phase 1 LSGD constructs \MG{localized} disjoint partitions (\ref{fig:tri0_sub1}--\ref{fig:tri0_sub4} and \ref{fig:tri2_sub1}--\ref{fig:tri2_sub4}) and Phase 2 LSGD produces an accurate approximation.}
	\label{fig:tsLSGD_tri}
\end{figure*}
Both neural networks are trained on the same data, $\{x_i,\mathbf{y}(x_i;p)\}_{i=1}^{n_{\text{data}}}$, where $x_i$ are uniformly sampled from [0,1] and $n_{\text{data}}=2000$. The standard ResNets are trained using the gradient descent method and the POUnets are trained using the LSGD method with $\lambda=0$ (Algorithm~\ref{thealg1}). The initial learning rate for Adam is set as $10^{-3}$. The maximum number of epochs $n_{\text{epoch}}$ is set as 2000 and we choose the neural network weights and biases that yield the best result on the training loss during the specified epochs.

Figure~\ref{fig:regression_errors} reports the relative $\ell^2$-errors of approximants produced by the standard ResNets and POUnets for varying $N_{\text{part}}$ and width for the piecewise linear functions (left) and the quadratic piecewise quadratic functions (right). The statistics are obtained from five independent runs. Figure~\ref{fig:regression_errors} essentially shows that the POUnets outperform the standard ResNets in terms of approximation accuracy; specifically, the standard ResNets with the ReLU activation function significantly fails to produce accurate approximantions, while POUnets obtain $<1\%$ error for large numbers of discontinuities.

Figure~\ref{fig:regression_snapshots} illustrates snapshots of the target functions and approximants produced by the two neural networks for target functions with the frequency parameter $p=\{4,5\}$. Figure~\ref{fig:regression_snapshots} confirms the trends shown in Figure \ref{fig:regression_errors}; the benefits of using POUnets are more pronounced in approximating functions with larger $p$ and in approximating quadratic functions (potentially, in approximating high-order polynomials). Figures \ref{fig:regression_snapshots_sub3}--\ref{fig:regression_snapshots_sub6} \MG{clearly} show that the POUnets accurately approximate the target functions with sub 1\% errors, while the standard ResNets significantly fails to produce accurate approximations.

\subsection{Results of two-phase LSGD}
Now we demonstrate effectiveness of the two-phase LSGD method (Algorithm~\ref{thealg2}) in constructing \MG{partitions which are localized according to the features in the data and nearly disjoint, i.e., breakpoints in the target function,} which \MG{we found} leads to better approximation accuracy. 

The first phase of the algorithm aims to construct \MG{such} partitions. To this end, we limit the expressibility of the polynomial regression by regularizing the Frobenius norm of the coefficients $\|\mathbf{c}\|_F^2$ in least-squares solves. This prevents a small number of partitions dominating and other partitions being collapsed \MG{per our discussion of the fast optimizer above}. These "quasi-uniform" partition are then used as the initial guess for the unregularized second phase. Finally, we study the qualitative differences in the resulting POU, particularly the ability of the network to learn a disjoint partition of space. We do however stress that there is no particular "correct" form for the resulting POU - this is primarily of interest because it shows the network may recover a traditional discontinuous polynomial approximation.

In the first phase, we employ a relatively larger learning rate in order to prevent weights and biases from getting stuck in local minima. On the other hand, in the second phase we employ a relatively smaller learning rate to ensure that the training error decreases. 

\begin{figure*}[t]
    \subfigure[Phase 1 (0)]{\includegraphics[height=28.5mm,width=28.5mm]{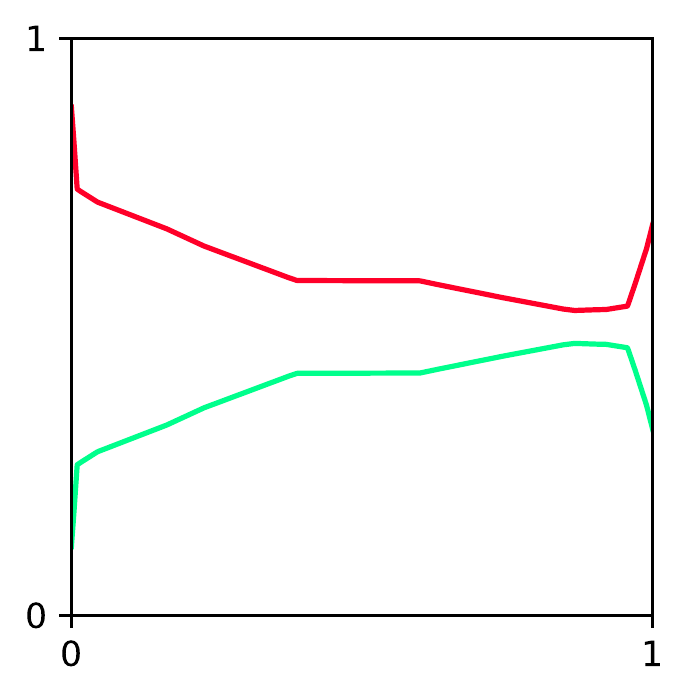}\label{fig:quad0_sub1}}
    \subfigure[Phase 1 (3000)]{\includegraphics[height=28.5mm,width=28.5mm]{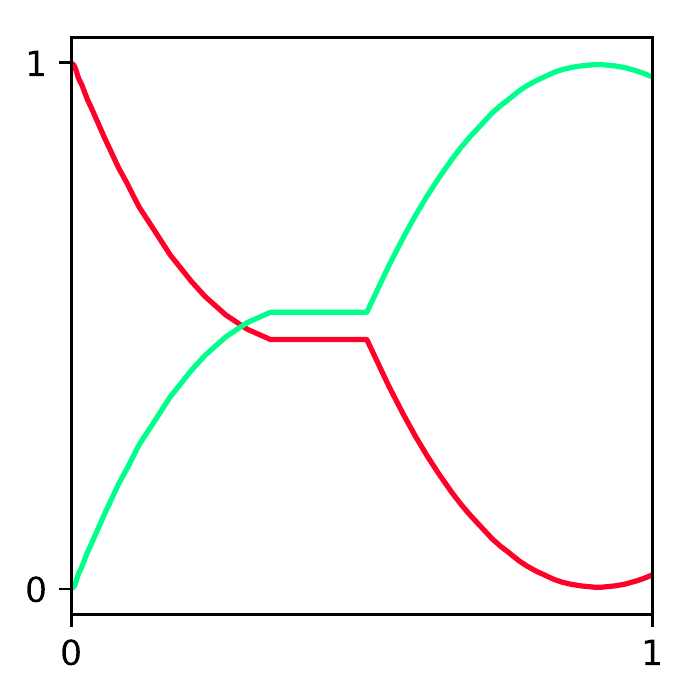}\label{fig:quad0_sub2}}
    \subfigure[Phase 1 (6000)]{\includegraphics[height=28.5mm,width=28.5mm]{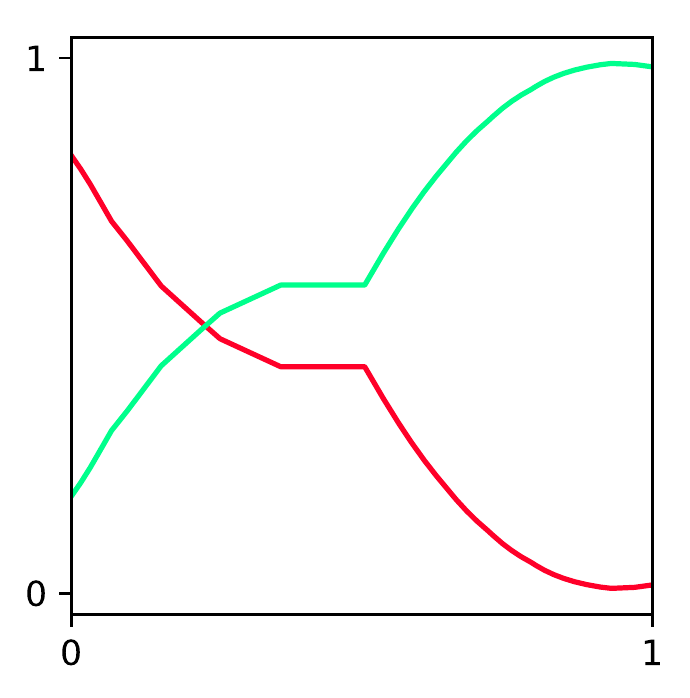}\label{fig:quad0_sub3}}
    \subfigure[Phase 1 (9000)]{\includegraphics[height=28.5mm,width=28.5mm]{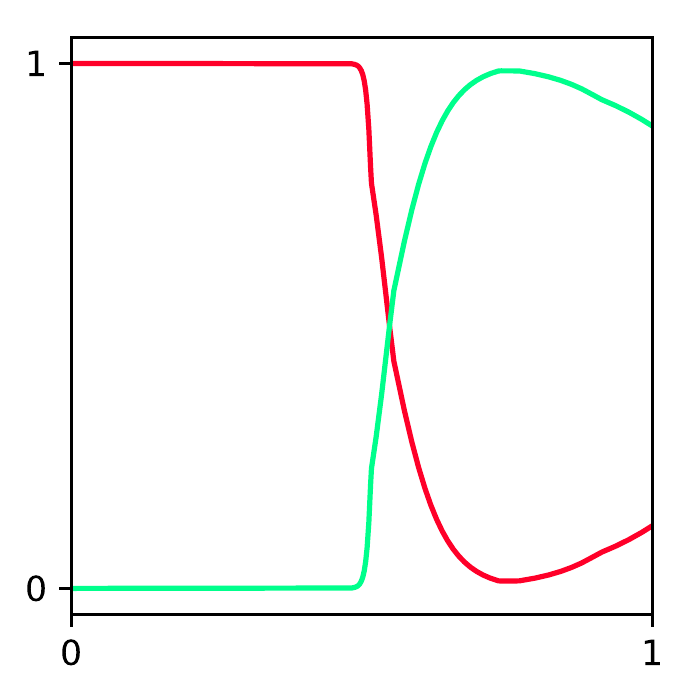}\label{fig:quad0_sub4}}
    \subfigure[Phase 2 (1000)]{
    \includegraphics[height=28.5mm,width=28.5mm]{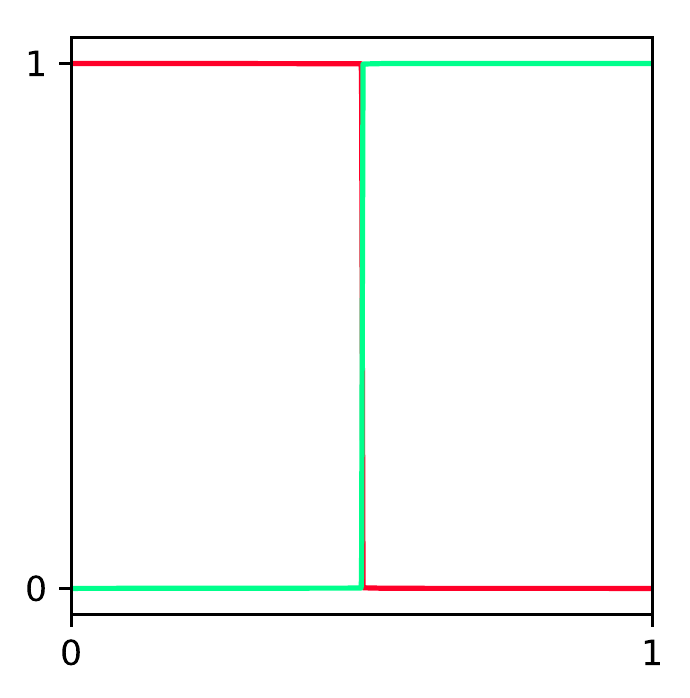}\label{fig:quad0_sub5}}
    \subfigure[Approximation]{\includegraphics[height=28.5mm,width=28.5mm]{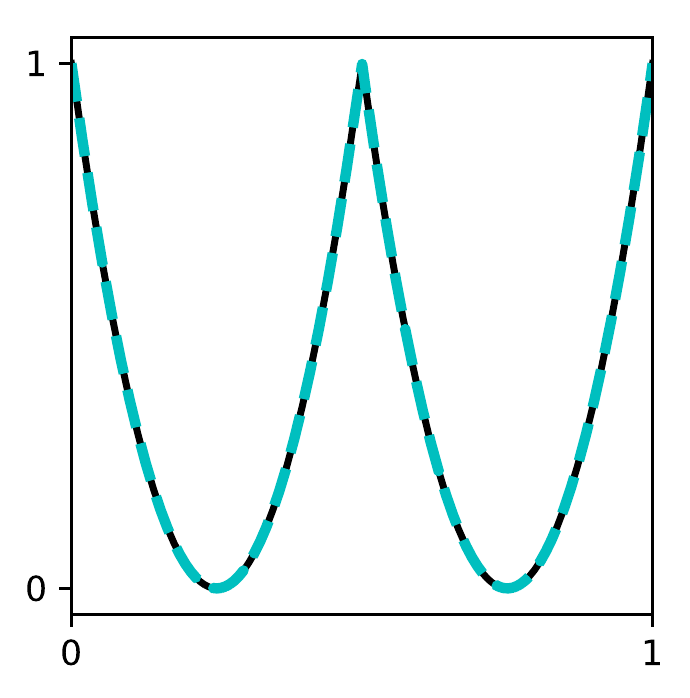}\label{fig:quad0_sub6}}\\

    \subfigure[Phase 1 (0)]{\includegraphics[height=28.5mm,width=28.5mm]{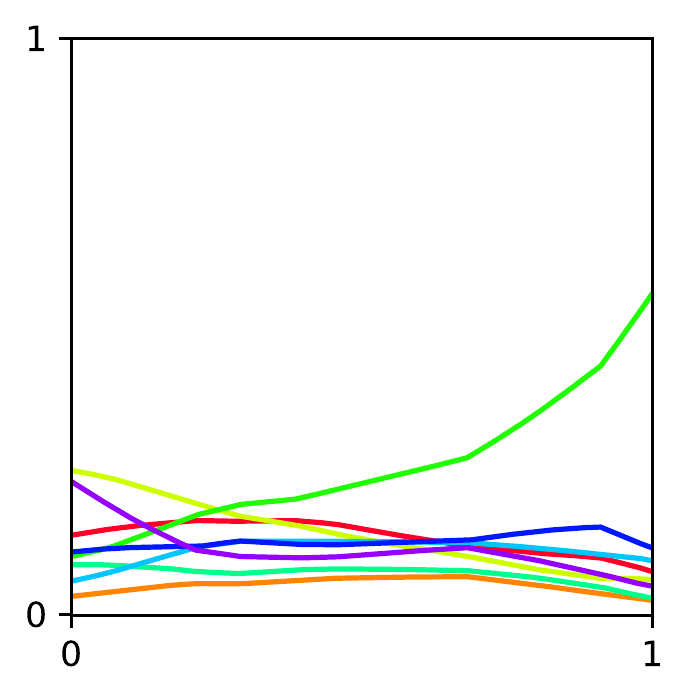}\label{fig:quad2_sub1}}
    \subfigure[Phase 1 (100000)]{\includegraphics[height=28.5mm,width=28.5mm]{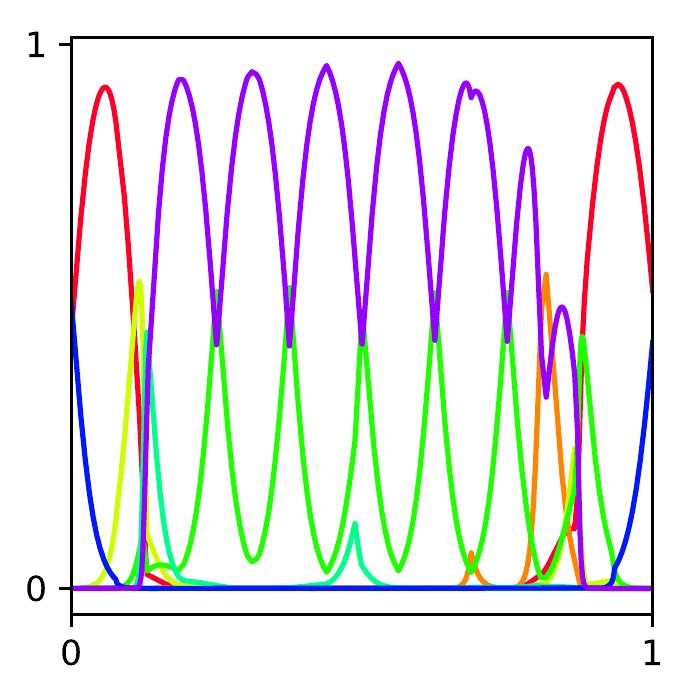}\label{fig:quad2_sub2}}
    \subfigure[Phase 1 (150000)]{\includegraphics[height=28.5mm,width=28.5mm]{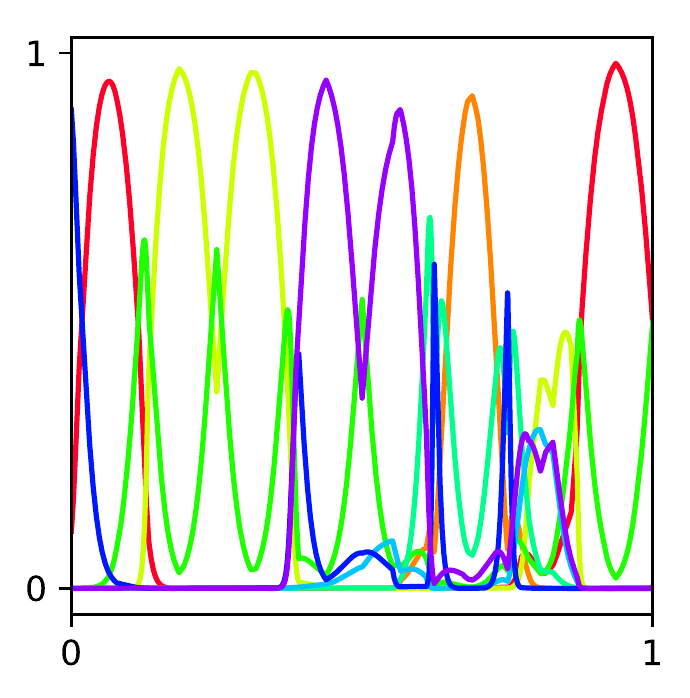}\label{fig:quad2_sub3}}
    \subfigure[Phase 1 (300000)]{\includegraphics[height=28.5mm,width=28.5mm]{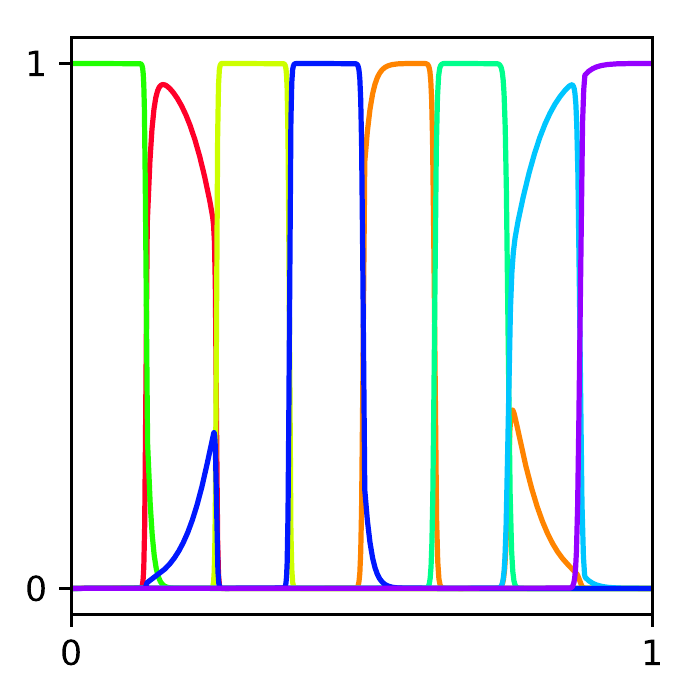}\label{fig:quad2_sub4}}
    \subfigure[Phase 1 (500000)]{
    \includegraphics[height=28.5mm,width=28.5mm]{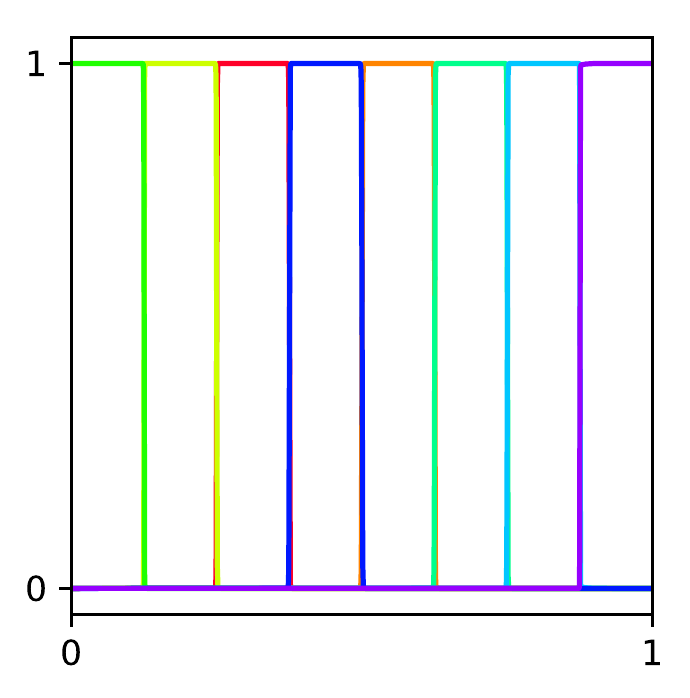}\label{fig:quad2_sub5}}
    \subfigure[Approximation]{\includegraphics[height=28.5mm,width=28.5mm]{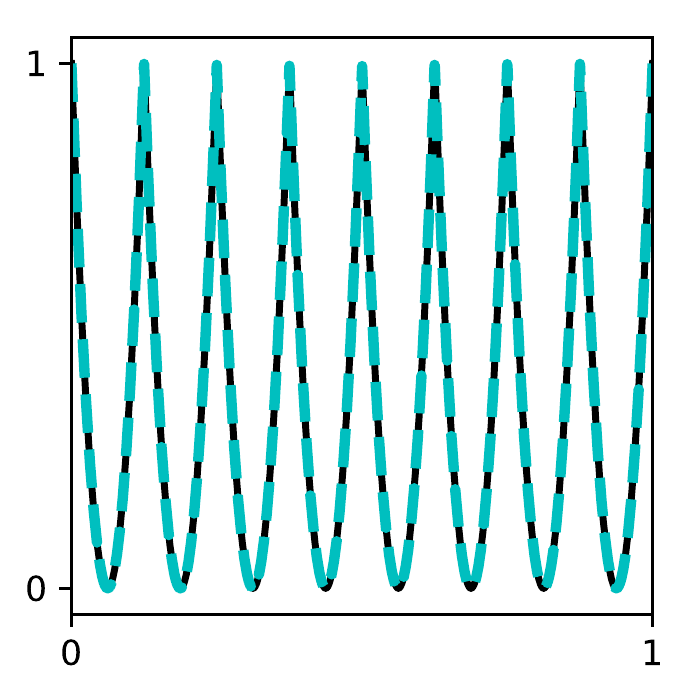}\label{fig:quad2_sub6}}
    \caption{Quadratic wave with two pieces (top) and quadratic wave with eight pieces (bottom): Phase 1 LSGD constructs \MG{localized} disjoint partitions (\ref{fig:quad0_sub1}--\ref{fig:quad0_sub4} and \ref{fig:quad2_sub1}--\ref{fig:quad2_sub4}) and Phase 2 LSGD produces an accurate approximation.}
	\label{fig:tsLSGD_quad2}
\end{figure*}

\paragraph{Triangular waves.} We demonstrate how the two-phase LSGD works in two example cases. The first set of example problems is the triangular wave with the frequency parameter $p=1,3$. We use the same POU network architecture described in the previous section (i.e., ResNet with width 8 and depth 8) and the same initialization scheme (i.e., the box initialization). We set 0.1 and 0.05 for the initial learning rates for phase 1 and the phase 2, respectively; we set \MG{the} other LSGD parameters as $\lambda=0.1$, $\rho=0.9$, and $n_{\text{stag}}=1000$. Figure \ref{fig:tsLSGD_tri} (top) depicts both how partitions are constructed during phase 1 (Figures~\ref{fig:tri0_sub1}--\ref{fig:tri0_sub4}), and snapshots of the partitions and the approximant at 1000th epoch of the phase 2 in Figures~\ref{fig:tri0_sub5} and \ref{fig:tri0_sub6}. The approximation accuracy measured in the relative $\ell^2$-norm of the error reaches down to $6.2042\times 10^{-8}$ after 12000 epochs in phase 2. 

Next we consider the triangular wave with the frequency parameter $p=3$. We use the POU network architecture constructed as ResNet with width 8 and dept 10, and use the box initialization. We set 0.05 and 0.01 for the initial learning rates for phase 1 and phase 2, respectively; we set \MG{the} other LSGD parameters as $\lambda=0.1$, $\rho=0.999$, and $n_{\text{stag}}=1000$. Figure \ref{fig:tsLSGD_tri} (bottom) depicts again how partitions evolve during the phase 1 and the resulting approximants. Figures \ref{fig:tri2_sub1}--\ref{fig:tri2_sub4} depicts that the two-phase LSGD constructs \MG{partitions that are disjoint and localized according to the features}
during the first phase.

\paragraph{Quadratic waves.} \MG{Finally, we approximate} the piecewise \MG{quadratic} wave with frequency parameter $p=1,3$ while employing the same network architecture used for approximating the triangular waves. For the $p=1$ case the learning rate set as 0.5 and 0.25 for phase 1 and phase 2. We use the same parameter settings (i.e.,  $\lambda=0.1$, $\rho=0.9$, and $n_{\text{stag}}=1000$) as in the previous experiment. Again, in phase 1 disjoint partitions are constructed, and the accurate approximation is produced in the phase 2 (Figures \ref{fig:quad0_sub1}--\ref{fig:quad0_sub6}). Moreover, we observe that the partitions are further refined during phase 2. For the $p=3$ case we again employ the architecture used for the triangular wave with $p=3$ (i.e., ResNet with width 8 and depth 10). We use the same hyperparameters as in the triangular wave with $p=3$ (i.e., 0.05 and 0.01 for learning rates, and $n_{\text{stag}}=1000$). The two exceptions are the weight for the regularization, $\lambda=1$ and $\rho=0.999$. Figures \ref{fig:quad2_sub1}--\ref{fig:quad2_sub5} illustrates that the phase 1 LSGD constructs disjoint supports for partitions, but takes much more epochs. Again, the two-stage LSGD produces an accurate approximation (Figure \ref{fig:quad2_sub6}).

\paragraph{Discussion.} The results of this section have demonstrated that it is important to apply a good regularizer during a pre-training stage to obtain approximately disjoint piecewise constant partitions. For the piecewise polynomial functions considered here, such partitions allowed in some cases recovery to near-machine precision. Given the abstract error analysis, it is clear that such \MG{localized} partitions \MG{which adapt to the features of the target function act similarly} to traditional \MG{$hp$}-approximation. There are several challenges regarding this approach, however: the strong regularization during phase 1 requires a large number of training epochs, and we were unable to obtain a set of hyperparameters which provide such clean partitions across all cases. This suggests two potential areas of future work. Firstly, an improved means of regularizing during pretraining that does not compromise accuracy may be able to train faster; \MG{deep learning strategies for parametrizing charts as in \cite{schonsheck2019chart} may be useful for this.} Secondly, it may be fruitful to understand the approximation error under less restrictive assumptions that the partitions are \MG{compactly supported or highly localized}. \MG{We have been guided by the idea that} polynomial\MG{s} defined on indicator function partitions reproduce the constructions used by the finite element community\MG{, but a less restrictive paradigm combined with an effective learning strategy may prove more flexible for general datasets.}

\section{Conclusions}
 \MG{Borrowing ideas from numerical analysis,} we have demonstrated an \MG{novel} architecture and optimization strategy \MG{the computational complexity of which need not scale} exponentially \MG{with the ambient} dimension and \MG{which provides} high-order convergence for smooth functions and error consistently under $1\%$ for \MG{piecewise smooth} problems. Such an architecture has the potential to provide DNN methods for solving high-dimensional PDE that converge in a manner competitive with traditional finite element spaces.

\section{Acknowledgements}

Sandia National Laboratories is a multimission laboratory managed and operated by National Technology and Engineering Solutions of Sandia, LLC, a wholly owned subsidiary of Honeywell International, Inc., for the U.S. Department of Energy’s National Nuclear Security Administration under contract {DE-NA0003530}.  This paper describes objective technical results and analysis.  Any subjective views or opinions that might be expressed in the paper do not necessarily represent the views of the U.S. Department of Energy or the United States Government. SAND Number: {SAND2020-6022 J}. 

The work of M. Gulian, R. Patel, and N. Trask are supported by the U.S. Department of Energy, Office of Advanced Scientific Computing Research under the Collaboratory on Mathematics and Physics-Informed Learning Machines for Multiscale and Multiphysics Problems (PhILMs) project. E. C. Cyr and N. Trask are supported by the Department of Energy early career program. M. Gulian is supported by the John von Neumann fellowship at Sandia National Laboratories.

\bibliography{SNL_AAAI}

\end{document}